\documentclass[10pt,twocolumn,letterpaper]{article}

\usepackage{iccv}
\usepackage{times}
\usepackage{epsfig}
\usepackage{graphicx}
\usepackage{amsmath}
\usepackage{amssymb}
\usepackage{multirow}
\usepackage{appendix}

\usepackage[pagebackref=true,breaklinks=true,letterpaper=true,colorlinks,bookmarks=false]{hyperref}

\iccvfinalcopy 

\def\iccvPaperID{****} 

\ificcvfinal\fi

\begin{document}

\title{Dynamic Clustering and Cluster Contrastive Learning for Unsupervised Person Re-identification}

\author{Ziqi He, Mengjia Xue, Yunhao Du, Zhicheng Zhao\thanks{Corresponding author}, Fei Su\\
Beijing University of Posts and Telecommunications, Beijing, China\\
{\tt\small \{heziqi, xuemengjia, dyh\_bupt, zhaozc, sufei\}@bupt.edu.cn}
}

\maketitle
\ificcvfinal\thispagestyle{empty}\fi

\begin{abstract}
Unsupervised Re-ID methods aim at learning robust and discriminative features from unlabeled data. However, existing methods often ignore the relationship between module parameters of Re-ID framework and feature distributions, which may lead to feature misalignment and hinder the model performance. To address this problem, we propose a dynamic clustering and cluster contrastive learning (DCCC) method. Specifically, we first design a dynamic clustering parameters scheduler (DCPS) which adjust the hyper-parameter of clustering to fit the variation of intra- and inter-class distances. Then, a dynamic cluster contrastive learning (DyCL) method is designed to match the cluster representation vectors' weights with the local feature association. Finally, a label smoothing soft contrastive loss ($L_{ss}$) is built to keep the balance between cluster contrastive learning and self-supervised learning with low computational consumption and high computational efficiency. Experiments on several  widely used public datasets validate the effectiveness of our proposed DCCC which outperforms previous state-of-the-art methods by achieving the best performance. Code is available at \url{https://github.com/theziqi/DCCC}.
\end{abstract}

\section{Introduction}

Person Re-identification (Re-ID) aims to identify the desired target pedestrian from a large number of cross-camera surveillance images. Initially, researchers have mostly worked on supervised Re-ID methods based on deep network models\cite{yi2014deep,li2014deepreid}. However, as supervised Re-ID methods are widely arranged in real-world scenarios\cite{ye2021deep}, the volume of data becomes larger and the time costs of manual annotation become more expensive, leading to limitations in the development of supervised Re-ID. Therefore, in order to reduce the cost of manual annotation, unsupervised Re-ID methods, which use unlabeled data for training, have received a lot of attention from researchers.

Unsupervised domain adaptation (UDA) Re-ID methods and fully unsupervised learning (USL) Re-ID methods are two types of unsupervised re-identification techniques. UDA Re-ID mthods involve a source domain without any annotation information and a target domain fully annotated based on transfer learning. However, the introduction of the source domain limits the model's performance. On one hand, the model's performance in the target domain is influenced by the quality of the knowledge learned in the source domain\cite{chen2021ice}. On the other hand, the discrepancy in data distribution between the source and target domains hinders sufficient knowledge transfer, which in turn inhibits the model's performance\cite{dai2022cluster}. USL Re-ID methods, in contrast, only employ unlabeled datasets for training, which are more adaptable and scalable without other external factors.

\begin{figure}[t]
\centering
   \includegraphics[width=\linewidth]{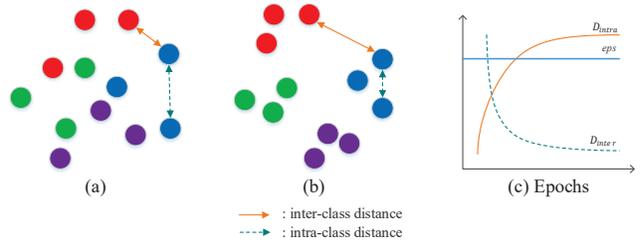}
   \caption{Visualization of inter-class and intra-class distances during training. (a) At the beginning, the inter-class distance is small and the intra-class distance is large, the density of the clusters is small and the discriminativeness of the feature vectors is relatively low. (b) After several training epochs, the inter-class distances become larger, the intra-class distances become smaller and the density of clusters becomes larger. (c) The clustering parameter $eps$ of DBSCAN is not fitting with the variation of distances in training\cite{dai2022cluster}.}
\label{fig:intra-inter}
\end{figure}

Recent methods commonly utilize plug-and-play modules, including a clustering algorithm, a memory bank and contrastive loss functions, together with network models to build an efficient and available USL Re-ID framework. Its remarkable performance has attracted a great deal of attention and taken a dominant place. They mostly follow this procedure in USL Re-ID task: (1) generating the corresponding pseudo-labels by a clustering algorithm; (2)  computing the contrastive loss by the supervision of pseudo-labels for the input instances, called query instances, with the postive and negative from the memory bank; (3) updating the cluster representation vectors in the memory bank for the next iteration. Each iteration stage drives the network to learn robust and discriminative features. In this case, the feature distribution converges towards the ground truth distribution. However, these methods always ignore this actual variation in defining each module. For instance, the most usually applied clustering algorithm in USL Re-ID method is DBSCAN: Cheng\cite{cheng2022hybrid} \etal assign different values to the clustering parameter $eps$ for different datasets, such as 0.5 on Market1501 and 0.6 on DukeMTMC-reID, while the values are 0.4 and 0.7 in ISE\cite{zhang2022implicit}. In general, the clustering parameters are usually taken empirically as the optimal fixed values\cite{wu2018unsupervised,ge2020self,dai2022cluster}. However, this approach of setting clustering parameters may result in a mismatch between the parameters and the feature distribution, called feature misalignment. On one side, as Figure \ref{fig:intra-inter} shows, during the training stage, the intra-class distances decrease, while the inter-class distances increase, and accordingly, the density of each cluster rises. On the other side, the clustering parameter $eps$ of DBSCAN is a distance threshold determining whether two neighbouring instances are of the same class which represents the density of the clusters. Thus, $eps$ is supposed to be adaptive to the distribution for prompting the clustering algorithm to generate high-quality pseudo-labels.

The observation inspires us to review the modules of the currently common USL Re-ID framework. We have found that feature misalignment also occurs in the memory bank and the contrastive loss functions. The cluster representation vectors is often the average centroid or the hardest instance in a mini-batch. However, they only reflect the local feature distribution. If the model learns the difference between them and the query features, it may cause a distribution shift. Moreover, contrastive methods do not cope well with the effects of distortion from data augmentation. Because of the traditional training way of a single network, after data augmentation, positive pseudo instances may be more different from query instances, negative pseudo instances may be more similar to query instances\cite{chen2021ice}, and the distribution of the same query features may be inconsistent.

To address the aforementioned problems, we propose a dynamic clustering and cluster contrastive learning (DCCC) method. DCCC designs a dynamic clustering parameters scheduler (DCPS) which continuously adjusts the clustering parameters to match the feature distribution as closely as possible during network training. We also propose a dynamic cluster contrastive learning (DyCL) method which optimizes the cluster representation vectors in the contrastive learning method by replacing the average centroid or the hardest positive instance with a dynamic cluster centroid. Based on the hard sample mining strategy\cite{shrivastava2016training}, the dynamic clustering centroid is the bridge between the weight parameters and the local feature distribution by assigning weights to each sample based on the feature distance in a mini-batch. In addition, DCCC also constructs a label smoothing soft contrastive loss ($L_{ss}$) function as the final loss function, which takes into account both cluster contrastive learning and self-supervised learning and reduces the computational cost.

To summarize, our contributions are: (1) We propose a dynamic clustering parameters scheduler, which enables the clustering hyper-parameters to dynamically decay with the training process. To our knowledge, our work is the first detailed study around the parameter settings of the clustering algorithm in unsupervised Re-ID research. (2) We propose a dynamic cluster contrastive learning method based on hard sample mining, which fully considers each instance in the mini-batch and assigns corresponding weights to them to update the cluter representation vectors in memory bank, solving the inconsistency problem. (3) We also use self-supervised methods combined with cluster contrast scheme to construct a simple and efficient label smooth soft contrastive loss function, which enhances the consistency in the distribution of same query features. (4) Experiments validate the effectiveness of our proposed method, which outperforms the state-of-the-art method in USL Re-ID on datasets such as Market1501 and DukeMTMC-reID.
\section{Related Works}

\subsection{Unsupervised Person Re-ID}

The most current unsupervised pedestrian Re-ID methods can be roughly divided into fully unsupervised (USL) methods and unsupervised domain adaptation (UDA) methods.

Due to the additional data introduced, UDA Re-ID often achieves better results than fully unsupervised methods. It requires labeled source domain data and unlabeled target domain data for training. Several UDA methods will transfer images from the source domain to the target domain utilizing Generative Adversarial Networks (GAN)\cite{wei2018person,zhong2018generalizing,chen2019instance,zou2020joint,chen2021joint}. The pseudo-label-based UDA methods will first pre-train the network on the source domain and then fine-tune the network on the target domain using the generated pseudo-labels\cite{song2020unsupervised,ge2020mutual,dai2021dual,zhai2020multiple,zhai2020ad,lin2020unsupervised}. Both of these methods focus on the transfer of knowledge from one domain to another. And USL Re-ID is more challenging and flexible in terms of data requirements as it only uses fully unlabeled data for training. Traditional methods use metric learning for personal retrieval tasks\cite{liao2015person,zheng2015scalable}. The performance bottleneck has been removed by clustering algorithms, and numerous USL Re-ID methods based on these algorithms have emerged\cite{lin2019bottom,zeng2020hierarchical}. Pseudo-labels generated by clustering algorithms or similarity estimation allow the model to train unlabeled data in a similar way to a labeled training model\cite{wang2020unsupervised,li2020joint,wu2019unsupervised,wang2020cycas}. SpCL\cite{ge2020self} is a  unsupervised Re-ID self-paced contrastive learning framework based on instance-level memory. Cluster Contrast\cite{dai2022cluster} employs cluster-level memory for addressing inconsistent updates of memory class centroids. ICE\cite{chen2021ice} incorporates the concepts of camera-aware, hard-sample mining, and soft-label in contrastive learning. However, the hard sample mining strategies in Re-ID's contrastive learning framework are often limited to a single hard sample and do not fully exploit global information.

\subsection{Clustering Algorithm}

A clustering algorithm's fundamental idea is dividing a dataset into clusters based on some criterion (typically distance) without any annotations. Clustering algorithms are frequently used in deep unsupervised learning due to their unsupervised capability and good performance. Clustering algorithms often calculate the similarity between samples using metrics \eg Euclidean distance, Manhattan distance, and Jaccard similarity coefficient\cite{hermans2017defense}. Clustering algorithms can be broadly categorised into Partition-based Methods, Density-based methods, Hierarchical Methods and so on. K-Means\cite{macqueen1965some}, a partition-based method, is first used in unsupervised Re-ID, but it requires the number of clusters as a hyper-parameter, which is difficult to estimate for unsupervised learning. DBSCAN\cite{ester1996density}, a density-based method, require the definition of two parameters $eps$ and $min\_samples$, which denote the neighbourhood radius of the density and the neighbourhood density threshold. The property of not requiring the number of clusters has led many unsupervised Re-ID algorithms to adopt DBSCAN. Nevertheless, fixed parameters of DBSCAN which is popular in cluster-based USL Re-ID method may lose the key context information at each training epoch.

\begin{figure*}
\centering
    \includegraphics[width=\linewidth]{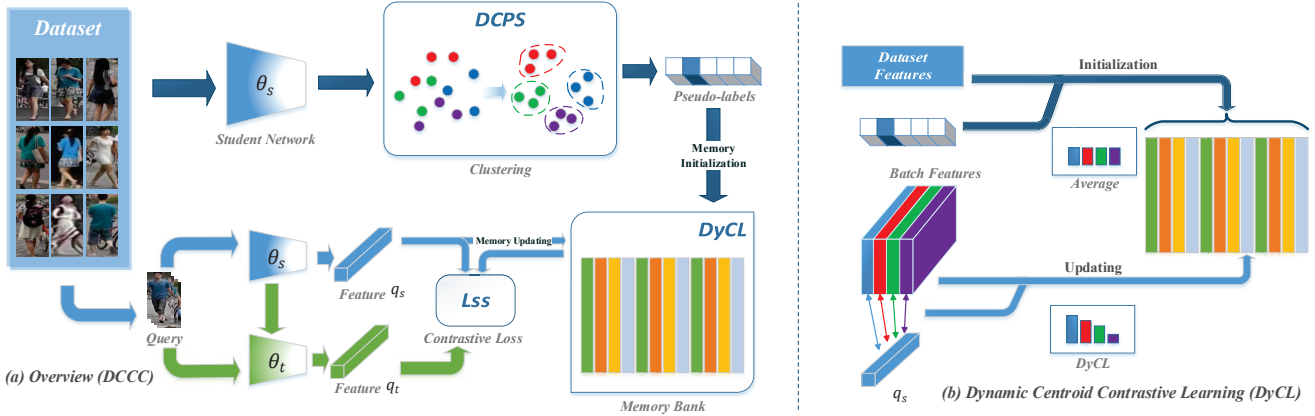}
   \caption{(a) The overview of our proposed \textbf{DCCC} framework. The top half shows the initialization stage of the memory bank, where the features are extracted by the student networks provided for the clustering algorithm with \textbf{DCPS} to generate the pseudo-labels. The bottom half shows the training stage, where the query instances are used to calculate $L_{ss}$. (b) \textbf{DyCL} updates the memory bank with dynamic centroid computed by the distances of the query instance and batch features in the same class, while the averages of overall dataset features are used to initialize.}
\label{fig:overview}
\end{figure*}

\section{Proposed Approach}

\subsection{Overview}
\label{sec:overview}

We propose a USL Re-ID framework that combines dynamic clustering algorithms and cluster contrastive learning. The framework is based on a general framework for self-supervised learning\cite{he2020momentum} and Cluster Contrast\cite{dai2022cluster}. Figure \ref{fig:overview} shows an overview of the framework. It includes backbone networks, a clustering algorithm, a memory bank, and other components. The overall training procedure is as follows.

In USL Re-ID methods, a dataset is commonly denoted as $X=\{x_1,x_2,...,x_N\}$, where each $x_i$ is an unlabeled image, containing $N$ pedestrian images totally. 

In each epoch of training, the initial stage is carried out first. The backbone networks consists of a teacher network and a student network. The student network is a regular network updated by gradient back-propagation, and its network parameters are denoted by $\theta_s$. Given a picture $x$, $f_{\theta_s}(x)$ represents the output features of the student network. The teacher network's parameters $\theta_t$ are updated by the parameters of the student network by exponential moving average (EMA)\cite{tarvainen2017mean}, which is formulated as follows:

\begin{equation}\label{eq:EMA}
\theta_t=\lambda\theta_t^{'}\ +\ (1-\lambda)\theta_s
\end{equation}

\noindent
where $\theta_t^{'}$ is the teacher network parameters from the previous iteration and $\lambda$ is a momentum update hyper-parameter. DBSCAN then divides the output features of the student network into sets of instances that can be clustered and outliers that are not clustered, based on Jaccard similarity coefficient\cite{hermans2017defense} pairs, and assigns them the corresponding pseudo-labels. The parameters of the clustering algorithm are determined by the dynamic clustering parameter scheduler (DCPS) that matches the global distribution of features. The mean value of the features in each instance set will be used as the initial cluster representation vector in the memory bank.

In the training stage, we use PK sampling\cite{jaccard1912distribution} where $P$ pedestrian IDs are randomly selected and $K$ pedestrian images are randomly drawn for each ID. Subsequently, the same query instances are fed into the student network and the teacher network after two different data augmentations. The dynamic custering contrastive learning (DyCL) performs a momentum update of the cluster representation vectors utilizing a hard sample mining strategy by the correlation between the student output features and the corresponding cluster representation vectors.

Eventually, our proposed label smoothing soft contrastive Loss ($L_{ss}$) is the final loss function, calculated from the output features and the cluster representation vectors.

\subsection{Dynamic Clustering Parameter Scheduler}
\label{sec:dcps}

The hyper-parameters of the clustering algorithms in USL Re-ID methods control the classification efficiency, \eg $eps$ in DBSCAN. However, in machine learning task, especially unsupervised Re-ID\cite{jin2020global,li2020joint,wang2022attentive}, these hyper-parameters are static, which does not match the features that are constantly changing during the training process. The density of the clusters and the value of $eps$ are closely related: if $eps$ is too large, there will be too much noise in the clusters, and if $eps$ is too small, many valid samples will be excluded. 

Coinciding with this paper, the learning rate\cite{patterson2017deep} which is the most fundamental hyper-parameter in neural network dynamically decays too. Motivated by learning rate, a dynamic $eps$ parameter strategies are given.

We strive to ensure that the $eps$ parameter variation curves fits the inter-class and intra-class distance variation curves in order to further align them. Then the inter- and intra-class distance variation curves are convex functions, which is demonstrated in Figure \ref{fig:intra-inter}. We thus choose the monotonical exponential function. The $eps$ value $\epsilon$ increases with each epoch to become a multiple of $\sigma_e$ from the previous epoch. It can be formulated as follows: 

\begin{equation}\label{eq:expoes}
\epsilon = \epsilon_{beign}*\sigma_e^{epochs}
\end{equation}
\noindent
where $\sigma_e\in[0,1]$  is the decay ratio, and $\epsilon_{beign}$ is the initial $eps$ value, witch is generally taken to be a larger number such as 0.7. Then $epochs$ is the current number of epochs in training.

Figure \ref{fig:dycl} shows function plots for different dynamic clustering parameter scheduler where our proposed exponential dynamic parameter scheduler converges more closely to the feature distribution than the other two. In Figure \ref{fig:intra-inter}, it can be seen that the $eps$ parameter continues to decay without end, while the curves of inter-and intra-class distance first fall and rise steeply, and finally level off. If $eps$ decreases when the feature distance is stable in the late training period, some information may be lost, which affects the quality of pseudo-labels. Therefore, we make the $eps$ parameter terminate when reaching a certain value (one half of the initial value).

\begin{figure}[t]
\centering
   \includegraphics[width=\linewidth]{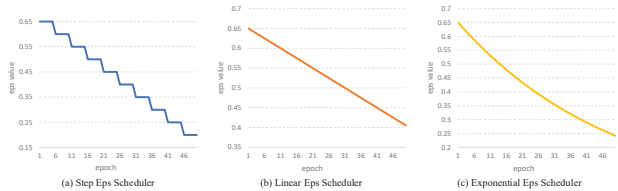}
   \caption{Different types of Dynamic EPS Scheduler.}
\label{fig:dcps}
\end{figure}

\subsection{Dynamic Cluster Contrastive Learning}
\label{sec:dycl}

Classical USL Re-ID methods based on contrastive learning calculate the loss in the context of a mini-batch. In order to break the constraint which the methods pick positive and negative samples locally, the features of SpCL\cite{ge2020self} are stored in the global memory and updated gradually with the training process. However, the batch training approach allows only some of the instances in a class to be updated in each iteration, leading to the unbalanced updating pace, which will shift the feature distribution. To solve this problem, Cluster Contrast\cite{dai2022cluster} stores the cluster representation vectors directly in the memory bank and updates them in a uniform momentum manner. Where the ClusterNCE loss\cite{dai2022cluster} can be formulated as:

\begin{equation}\label{eq:clusternceloss}
L_{cluster} = \mathbb{E}\left[-log\frac{exp(q \cdot c_+ / \tau)}{{\sum ^{C}_{i=1}}exp(q \cdot c_i/\tau)}\right]
\end{equation}

\noindent
where $c_i$ is the $i$-th cluster representation vector stored in memory bank, $c_+$ is the cluster representation vector corresponding to $q$, $\tau$ is the temperature factor and $C$ is the total number of clusters in pseudo-labels. Cluster representation vectors speed up the convergence of the model and also alleviate the inevitable drawbacks of the false positive due to the uncertainty of the clustering algorithm. Cluster representation vectors should reflect as much information about the class as possible to ensure learning accuracy. However, Cluster Contrast uses the average centroid or the hardest sample that do not reflect the holistic distribution.

\begin{figure}[t]
\centering
   \includegraphics[width=\linewidth]{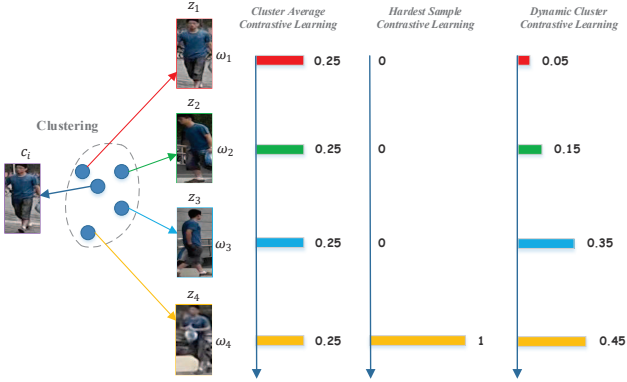}
   \caption{Different ways to assign weights of cluster representation vectors.}
\label{fig:dycl}
\end{figure}

Inspired by adaptive weight triplet loss\cite{ristani2018features}, we propose a dynamic clustering contrastive learning (DyCL) method. Contrary to adaptive weight triplet loss, which adopts the hard sample mining strategy in pairwise loss to obtain the distance between positive and negative pairs, we employ dynamic weights to memory momentum updating, so that the model can fully exploit the valid information in the global context. According to the hard sample mining strategy, we assign a corresponding weight to similar instances of each query instance, with the harder instances having a greater weight. A softmax function is used to gain the weights of the samples in order to emphasize the significance of hard instances and prevent entering a local optimum due by:

\begin{equation}\label{eq:dyclloss}
w_{ij}^{dy} = \frac{exp(-<c_i \cdot z_j>/\tau_{w})}{\sum_{m=1}^{N_i}exp(-<c_i \cdot z_m>/\tau_{w})}
\end{equation}

\noindent
where $\tau_w$ is the temperature coefficient hyper-parameter that affects the proportion of weights for hard instances, $N_i$ is the instance number of $i$-th class in a mini-batch and $z_m$ is the $m$-th instance feature of $N_i$. Note that the sum of weights is $\sum_{j=1}^{N_i}w_{ij}^{dy} = 1$. Thus, the $i$-th dynamic cluster centroid is the weighted mean in the mini-batch:

\begin{equation}\label{eq:weightmean}
\hat{c_i} = \sum_{j=1}^{N_i} w_{ij}z_j
\end{equation}

\begin{equation}\label{eq:commonupdate}
c_i \leftarrow \gamma{c_i} + (1-\gamma)\hat{c_i}
\end{equation}

\noindent
where $\gamma$ is the hyper-parameter of momentum updating. A comparison with the other two contrastive learning methods is shown in Figure \ref{fig:dycl}. It can be seen that DyCL is more responsive to the details of the feature distribution and that the adjustable $\tau_w$ hyper-parameter can further balance the global and local information.

\subsection{Label Smoothing Soft Contrastive Loss}
\label{sec:lss}

We have investigated the loss functions of existing USL Re-ID methods and found two potential improvements: \textbf{Firstly}, these methods tend to use more complex loss functions in order to enable the model to learn multiple tasks simultaneously. ICE\cite{chen2021ice} incorporates proxy loss, hard instance contrastive loss, and soft instance contrastive loss to reduce intra-class distribution differences and mitigate the distortion due to data augmentation. HDRCL\cite{cheng2022hybrid} combines pseudo-label-based local-to-global contrastive loss and self-supervised probabilistic regression loss to enable the model to generate more discriminative features. PPLR\cite{cho2022part} is even more targeted at the proposed part-based unsupervised Re-ID framework with four loss functions containing camera loss, local feature loss, global feature loss, and triplet loss. Besides the relatively heavier computational cost of loss and back-propagation, the overly complicated hyper-parameters and weights also make it difficult to tune. \textbf{Secondly}, since data augmentation may cause a degree of distortion, the features of the same query instances may differ, resulting in the similarity change within features, which affects the consistency of the feature distribution and possibly undermines the model performance. MMT\cite{ge2020mutual} proposes a soft classification loss based on the 'Teacher-Student' model\cite{tarvainen2017mean}, where computes the temporally averaged similarity probability of the teacher network output as a soft label to supervise the the student network, and the output of both networks can be consistent. This self-supervised method avoids the error amplification from perturbations in the feature distribution during training in a non-parametric manner\cite{cheng2022hybrid}.

To achieve the above goals, we try to combine contrast learning and self-supervised learning with only one loss function. We propose a label smoothing soft contrastive loss ($L_{ss}$) based on pseudo-labels refinement for the prediction of teacher features and clustering results. Specifically, we generate a refined smoothing soft label $y^{sm}_k$ for each query instance:

\begin{equation}\label{eq:smoothsoftlabel}
y^{sm}_k=\mu_{s}y^t_k+(1-\mu_{s}) \tilde{y_k}
\end{equation}

\noindent
where $\tilde{y_k}$ represents the clustering-generated pseudo-label of the query instance which is an one-hot label. $\mu_s\in[0,1]$ is a weight parameter controlling the ratio of the soft label $y^t_k$ to the one-hot pseudo-label $\tilde{y_k}$. Soft label $y^t_k$  is the similarity probability of the query feature to the $k$-th class representation vector, which can be formulated as:

\begin{equation}\label{eq:yklabel}
y^t_k=\frac{exp(q_t \cdot c_k / \tau)}{{\sum ^{C}_{i=1}}exp(q_t \cdot c_i/\tau)}
\end{equation}
\noindent
where $q_t$ represents the query features from the teacher network, $c_i$ is the $i$-th cluster representation vector, and $\alpha$ is the temperature coefficient hyper-parameter. In contrast to the one-hot pseudo-label which only exploits class information, the refined smoothing soft label additionally considers the consistency of feature distribution of the same query instance. Since ClusterNCE loss is effectively a cross-entropy loss\cite{he2020momentum}, plugging $y^{sm}_k$ into Eq. \ref{eq:clusternceloss}, we can obtain the formula for the label smoothing soft contrastive loss as:

\begin{equation}\label{eq:smoothsoftloss}
L_{ss}=\sum_{k=1}^{C} -{y}^{sm}_k \cdot log(y^s_k)
\end{equation}
\noindent
where $y^t_k$ denotes the similarity probability of the teacher network, which also has the same expression as Eq. \ref{eq:yklabel}. Unlike previous work with a loss function consisting of multiple parts, the label smoothing soft contrastive loss $L_{ss}$ integrates the contrastive and self-supervised methods with high computational efficiency and low computational consumption.

\begin{figure}[t]
\centering
   \includegraphics[width=\linewidth]{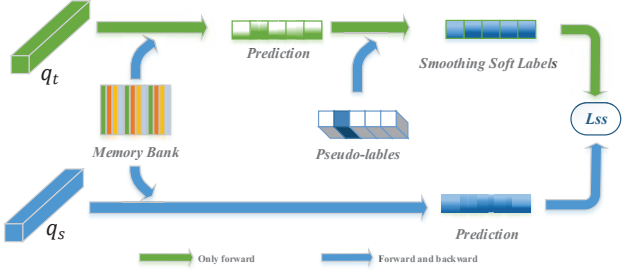}
   \caption{The detailed pipeline of label smoothing soft contrastive loss $L_{ss}$.}
\label{fig:dcps}
\end{figure}

\section{Experiment}

\subsection{Dataset and Evaluation Protocol}

We validate the proposed method on three generic person Re-ID datasets, Market1501\cite{zheng2015scalable}, DukeMTMC-reID\cite{ristani2016performance}, and MSMT17\cite{wei2018person}, respectively.

Market-1501 is collected from 6 cameras at the entrance of supermarkets in Tsinghua University campus. 12936 pedestrian images from 751 pedestrian IDs are included in the training set, and 19732 pedestrian images from 750 pedestrian IDs are included in the test set.

DukeMTMC-reID is a subset of the DukeMTMC dataset, which is derived from 8 non-overlapping cameras. 16522 images from 702 pedestrian IDs in the dataset are training images, 2228 images from another 702 pedestrian IDs are query images, and 17661 are gallery images.

MSMT17 is a recently proposed large person Re-ID dataset, containing a total of 126441 images from 4101 pedestrian IDs captured from 15 cameras, of which the training set contains 32621 images from 1041 pedestrian IDs and the test set contains 93820 images from 3060 pedestrian IDs.

Our experiments validate the model performance using Rank-1, Rank-5 and Rank-10 precision of cumulative matching characteristics (CMC)\cite{wang2007shape} and mean average precision (mAP)\cite{zheng2015scalable} metrics. We don't use any post-processing operations, such as re-ranking\cite{zhong2017re}.

\subsection{Implementation}

\noindent
\textbf{Network structure.} We used 4 Tesla V100 GPUs and Resnet50\cite{he2016deep} as the encoder backbone for feature extraction,  pre-trained on ImageNet\cite{deng2009imagenet}. We removed all modules after the fourth convolutional layer of the backbone networks and added a global average pooling layer (GAP), followed by a batch normalization layer\cite{ioffe2015batch} and a L2 normalization layer, with 2048-dimensional output features.

\medskip
\noindent
\textbf{Parameter settings.} We use a warm-up strategy, where the learning rate grows linearly to 0.00035 for the first 20 epochs, and then stops decaying. We use Adam with a weight decay factor of 5e-4. The network performs 200 iterations in one epoch, for a total of 70 epochs. Each mini-batch contains 64 images from 4 pedestrian IDs, which means the batch size is 256. Before being fed into the network, each image is scaled down to a pixel size of $256\times128$ with data augmentation, including random inverting, random cropping and random erasing. We set the momentum update parameters  $\gamma$ to 0.1 in Eq. \ref{eq:commonupdate}, $\tau$ of contrastive loss to 0.05 in Eq. \ref{eq:clusternceloss} and $\tau_w$ of dynamic weights to 0.09 in Eq. \ref{eq:dyclloss}. Before the start of each epoch, we calculate the feature distances by Jaccard coefficients with a nearest neighbour parameter $k$ of 30, and then DBSCAN generates the corresponding pseudo-labels for all features in the dataset, with the minimal number of neighbors set to 4.

\subsection{Ablation Study}

\begin{figure}[t]
\centering
   \includegraphics[width=\linewidth]{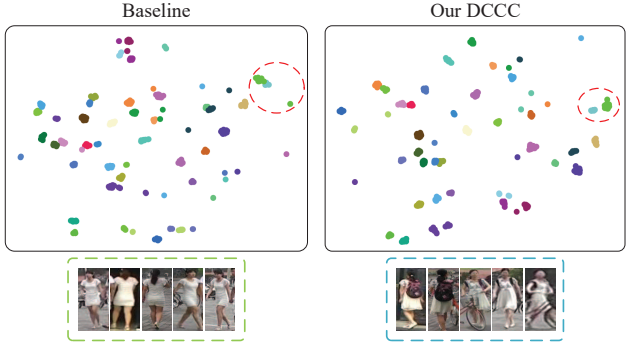}
   \caption{T-SNE visualization\cite{van2008visualizing} features extracted by the models of the baseline and our DCCC. Different colors represent different ground truth person IDs. Typical examples are marked by red dashed ellipses. }
\label{fig:tsne}
\end{figure}

To validate the effectiveness of our proposed method, we conducted detailed comparative experiments on Market1501 and DukeMTMC-reID.  We choose Cluster Contrast\cite{dai2022cluster} as the baseline method for our experiments, which updates the cluster representation vectors directly with query instances. We verify the effectiveness of our proposed three components: the dynamic clustering parameter scheduler (DCPS), the dynamic cluster contrastive learning (DyCL) and the label smoothing soft contrastive loss ($L_{ss}$), and the results are shown in Tab. \ref{tab:ablationstudy}. We intuitively visualize the features extracted by baseline model and our DCCC model utilizing T-SNE\cite{van2008visualizing} as shown in Figure \ref{fig:tsne} which shows our cluster compactness and independence.

\begin{table}[]
\centering
\resizebox{\linewidth}{!}{%
\begin{tabular}{ccc|c|cc|cc}
\hline
\multicolumn{3}{c|}{Components} & \multirow{2}{*}{No.} & \multicolumn{2}{c|}{Market1501} & \multicolumn{2}{c}{DukeMTMC-reID} \\ \cline{1-3} \cline{5-8} 
DCPS & DyCL & $L_{ss}$ &  & mAP & Rank-1 & mAP & Rank-1 \\ \hline
 &  &  & $\sharp 1$ & 80.9 & 91.7 & 71.5 & 84.4 \\
 & \checkmark &  & $\sharp 2$ & 83.1 & 92.7 & 71.9 & 83.7 \\
\checkmark &  &  & $\sharp 3$ & 81.7 & 92.0 & 72.3 & 84.4 \\
 &  & \checkmark & $\sharp 4$ & 84.8 & 93.4 & 72.5 & 84.5 \\
\checkmark & \multicolumn{1}{l}{} & \checkmark & $\sharp 5$ & 85.4 & 93.7 & 73.5 & 85.5 \\
\multicolumn{1}{l}{} & \checkmark & \checkmark & $\sharp 6$ & 86.3 & 94.7 & 73.1 & 84.6 \\
\checkmark & \checkmark &  & $\sharp 7$ & 83.3 & 93.3 & 72.0 & 84.0 \\
\checkmark & \checkmark & \checkmark & $\sharp 8$ & \textbf{86.6} & \textbf{94.1} & \textbf{74.0} & \textbf{85.4} \\ \hline
\end{tabular}%
}
\caption{The effectiveness of each components in our proposed dynamic clustering and cluster contrastive learning (DCCC). DCCC includes the dynamic clustering parameter scheduler (DCPS), the the dynamic cluster contrastive learning (DyCL) and the label smoothing soft contrastive loss ($L_{ss}$).}
\label{tab:ablationstudy}
\end{table}

\medskip
\noindent
    \textbf{Effectiveness of dynamic clustering parameter scheduler (DCPS).} The effectiveness of DCPS is demonstrated by the comparison between the results in ($\sharp 1$ and $\sharp 3$), ($\sharp 4$ and $\sharp 5$), and ($\sharp 6$ and $\sharp 8$), as shown in Tab.\ref{tab:ablationstudy}. DCPS enables the clustering parameter to no longer be fixed, but to vary dynamically with the feature distribution. The experimental results imply that DCPS benefits the clustering algorithm in generating high quality pseudo-labels. DCPS eventually uses the exponential EPS scheduler (ExpoES) from Sec. \ref{sec:dcps} for DBSCAN. To further validate the effectiveness of ExpoES, we compare it with the step EPS scheduler (StepES) and the linear EPS scheduler (LinearES). As shown in Tab. \ref{tab:dcps}. ExpoES has the most significant improvement on baseline, with mAP improving by 0.8 on Market1501 and DukeMTMC-reID.

\begin{table}[]
\centering
\resizebox{.8\linewidth}{!}{%
\begin{tabular}{cc|cc|cc}
\hline
\multicolumn{2}{c|}{\multirow{2}{*}{Methods}} & \multicolumn{2}{c|}{Market1501} & \multicolumn{2}{c}{DukeMTMC-reID} \\ \cline{3-6} 
\multicolumn{2}{c|}{}                                 & mAP           & Rank-1        & mAP           & Rank-1        \\ \hline
\multicolumn{2}{c|}{baseline}                         & 80.9          & 91.7          & 71.5          & 84.4          \\ \hline
\multicolumn{1}{c|}{\multirow{3}{*}{DCPS}} & StepES   & 81.3          & 92.2          & 71.8          & 84.5          \\
\multicolumn{1}{c|}{}                      & LinearES & 81.6          & \textbf{92.7} & 72.0          & \textbf{85.1} \\
\multicolumn{1}{c|}{}                      & \textbf{ExpoES}   & \textbf{81.7} & 92.0          & \textbf{72.3} & 84.4          \\ \hline
\end{tabular}%
}
\caption{Comparison with different dynamic clustering parameter schedulers.}
\label{tab:dcps}
\end{table}

\medskip
\noindent
\textbf{Effectiveness of dynamic cluster contrastive learning (DyCL).} As shown in Tab. \ref{tab:dycl}, the effectiveness of DyCL is illustrated by the comparison between the results in ($\sharp 1$ and $\sharp 2$), ($\sharp 4$ and $\sharp 6$) and ($\sharp 6$ and $\sharp 8$). DyCL enables cluster representation vectors to be aligned with the feature distribution locally, enabling the model to learn finer-grained knowledge. We also experiment with the two non-dynamic cluster contrastive learning methods in Sec. \ref{sec:dycl}. The results of the baseline, the cluster average contrastive learning (AvgCL) and the hardest sample contrastive learning (HardestCL) are not as good as the performance of DyCL in Tab. \ref{tab:dycl}.

\medskip
\noindent
\textbf{Effectiveness of label smoothing soft contrastive loss ($L_{ss}$).}  The efficiency of $L_{ss}$ can be observed in the comparison between ($\sharp 1$ and $\sharp 4$), ($\sharp 2$ and $\sharp 6$), ($\sharp 3$ and $\sharp 5$) and ($\sharp 7$ and $\sharp 8$) in Tab. \ref{tab:ablationstudy}. In terms of the performance, the baseline is greatly improved by 3.9\%/1.7\% and 1.0\%/0.1\% mAP/Rank-1 on Market1501 and DukeMTMC-reID via $L_{ss}$. In order to lessen the interference brought on by data augmentation, $L_{ss}$ is built on a dual network topology including probabilistic distillation and contrastive loss based on label refinement. Additionally, the comparison of $L_{ss}$ and cross-entropy loss can be seen in the appendix.
\begin{table}[]
\centering
\resizebox{.7\linewidth}{!}{%
\begin{tabular}{c|cc|cc}
\hline
\multirow{2}{*}{Methods} & \multicolumn{2}{c|}{Market1501} & \multicolumn{2}{c}{DukeMTMC-reID}  \\ \cline{2-5} 
                         & mAP            & Rank-1         & mAP           & Rank-1        \\ \hline
baseline\cite{dai2022cluster}                 & 80.9           & 91.7           & 71.5          & \textbf{84.4}          \\
AvgCL                    & 79.2           & 91.9           & 62.0          & 78.0          \\
HardestCL                & 82.4           & 92.3           & 70.4          & 83.4          \\
\textbf{DyCL}            & \textbf{83.1}  & \textbf{92.7}  & \textbf{71.9} & 83.7 \\ \hline
\end{tabular}%
}
\caption{Comparison of different contrastive learning methods of updating manner for cluster representation vectors.}
\label{tab:dycl}
\end{table}

\begin{table*}[]
\centering
\resizebox{.9\linewidth}{!}{%
\begin{tabular}{cccccccccccccc}
\hline
\multicolumn{1}{c|}{} & \multicolumn{1}{c|}{} & \multicolumn{4}{c|}{Market1501} & \multicolumn{4}{c|}{DukeMTMC-reID} & \multicolumn{4}{c}{MSMT17} \\ \cline{3-14} 
\multicolumn{1}{c|}{\multirow{-2}{*}{\textbf{Methods}}} & \multicolumn{1}{c|}{\multirow{-2}{*}{\textbf{Reference}}} & mAP & Rank-1 & Rank-5 & \multicolumn{1}{c|}{Rank-10} & mAP & Rank-1 & Rank-5 & \multicolumn{1}{c|}{Rank-10} & mAP & Rank-1 & Rank-5 & Rank-10 \\ \hline
\multicolumn{14}{l}{\textit{\textbf{Fully Unsupervised}}} \\
\multicolumn{1}{c|}{LOMO\cite{liao2015person}} & \multicolumn{1}{c|}{CVPR'15} & 8.0 & 27.2 & 41.6 & \multicolumn{1}{c|}{49.1} & 4.8 & 12.3 & 21.3 & \multicolumn{1}{c|}{26.6} & - & - & - & - \\
\multicolumn{1}{c|}{BOW\cite{zheng2015scalable}} & \multicolumn{1}{c|}{ICCV'15} & 14.8 & 35.8 & 52.4 & \multicolumn{1}{c|}{60.3} & 8.3 & 17.1 & 28.8 & \multicolumn{1}{c|}{34.9} & - & - & - & - \\
\multicolumn{1}{c|}{BUC\cite{lin2019bottom}} & \multicolumn{1}{c|}{AAAI'19} & 38.3 & 66.2 & 79.6 & \multicolumn{1}{c|}{84.5} & 27.5 & 47.4 & 62.6 & \multicolumn{1}{c|}{68.4} & - & - & - & - \\
\multicolumn{1}{c|}{MMCL\cite{wang2020unsupervised}} & \multicolumn{1}{c|}{CVPR'20} & 45.5 & 80.3 & 89.4 & \multicolumn{1}{c|}{92.3} & 51.4 & 72.4 & 82.9 & \multicolumn{1}{c|}{85.0} & 11.2 & 35.4 & 44.8 & 49.8 \\
\multicolumn{1}{c|}{JVTC\cite{li2020joint}} & \multicolumn{1}{c|}{ECCV'20} & 41.8 & 72.9 & 84.2 & \multicolumn{1}{c|}{88.7} & 42.2 & 67.6 & 78.0 & \multicolumn{1}{c|}{81.6} & 15.1 & 39.0 & 50.9 & 56.8 \\
\multicolumn{1}{c|}{JVTC+\cite{li2020joint}} & \multicolumn{1}{c|}{ECCV'20} & 47.5 & 79.5 & 89.2 & \multicolumn{1}{c|}{91.9} & 50.7 & 74.6 & 82.9 & \multicolumn{1}{c|}{85.3} & 17.3 & 43.1 & 53.8 & 59.4 \\
\multicolumn{1}{c|}{HCT\cite{zeng2020hierarchical}} & \multicolumn{1}{c|}{CVPR'20} & 56.4 & 80.0 & 91.6 & \multicolumn{1}{c|}{95.2} & 50.7 & 69.6 & 83.4 & \multicolumn{1}{c|}{87.4} & - & - & - & - \\
\multicolumn{1}{c|}{SpCL\cite{ge2020self}} & \multicolumn{1}{c|}{NeurIPS'20} & 73.1 & 88.1 & 95.1 & \multicolumn{1}{c|}{97.0} & 65.3 & 81.2 & 90.3 & \multicolumn{1}{c|}{92.2} & 19.1 & 42.3 & 55.6 & 61.2 \\
\multicolumn{1}{c|}{CAP$\dagger$\cite{wang2021camera}} & \multicolumn{1}{c|}{AAAI'21} & 79.2 & 91.4 & 96.3 & \multicolumn{1}{c|}{97.7} & 67.3 & 81.1 & 89.3 & \multicolumn{1}{c|}{91.8} & 36.9 & 67.4 & 78.0 & 81.4 \\
\multicolumn{1}{c|}{ICE$\dagger$\cite{chen2021ice}} & \multicolumn{1}{c|}{ICCV'21} & 82.3 & 93.8 & 97.6 & \multicolumn{1}{c|}{98.4} & 69.9 & 83.3 & 91.5 & \multicolumn{1}{c|}{94.1} & {38.9} & {70.2} & {80.5} & {84.4} \\
\multicolumn{1}{c|}{ICE(w/o camera)\cite{chen2021ice}} & \multicolumn{1}{c|}{ICCV'21} & 79.5 & 92.0 & 97.0 & \multicolumn{1}{c|}{98.1} & 67.2 & 81.3 & 90.1 & \multicolumn{1}{c|}{93.0} & 29.8 & 59.0 & 71.7 & 77.0 \\
\multicolumn{1}{c|}{Cluster Contrast\cite{dai2022cluster}} & \multicolumn{1}{c|}{ACCV'21} & 82.1 & 92.3 & 96.7 & \multicolumn{1}{c|}{97.9} & 72.6 & 84.9 & 91.9 & \multicolumn{1}{c|}{93.9} & 27.6 & 56.0 & 66.8 & 71.5 \\
\multicolumn{1}{c|}{IIDS$\dagger$\cite{xuan2022intra}} & \multicolumn{1}{c|}{TPAMI'22} & 78.0 & 91.2 & 96.2 & \multicolumn{1}{c|}{97.7} & 68.7 & 82.1 & 90.8 & \multicolumn{1}{c|}{93.7} & 35.1 & 64.4 & 76.2 & 80.5 \\
\multicolumn{1}{c|}{PPLR(w/o camera)} & \multicolumn{1}{c|}{CVPR'22} & 81.5 & 92.8 & 97.1 & \multicolumn{1}{c|}{98.1} & - & - & - & \multicolumn{1}{c|}{-} & 31.4 & 61.1 & 73.4 & 77.8 \\
\multicolumn{1}{c|}{PPLR$\dagger$} & \multicolumn{1}{c|}{CVPR'22} & 84.4 & \color{blue} 94.3 & 97.8 & \multicolumn{1}{c|}{98.6} & - & - & - & \multicolumn{1}{c|}{-} & \color{blue} 42.2 & \color{blue} 73.3 & \color{blue} 83.5 & \color{blue} 86.5 \\
\multicolumn{1}{c|}{\textbf{DCCC(Ours)}} & \multicolumn{1}{c|}{This paper} & \color{blue} 86.6 & 94.1 & \color{blue} 98.0 & \multicolumn{1}{c|}{\color{blue} 98.9} & \color{blue} 74.0 & \color{blue} 85.4 & \color{blue} 92.2 & \multicolumn{1}{c|}{\color{blue} 93.9} & \color{blue} 31.6 & \color{blue} 62.3 & \color{blue} 73.4 & \color{blue} 77.9 \\
\multicolumn{1}{c|}{\textbf{DCCC*}} & \multicolumn{1}{c|}{This paper} & {\color{red} 88.2} & {\color{red} 95.2} & {\color{red} 98.3} & \multicolumn{1}{c|}{{\color{red} 99.0}} & {\color{red} 76.9} & {\color{red} 87.1} & {\color{red} 93.5} & \multicolumn{1}{c|}{{\color{red} 95.2}} & {\color{red} 44.3} & {\color{red} 73.8} & {\color{red} 82.5} & {\color{red} 85.7} \\ \hline
\multicolumn{14}{l}{\textit{\textbf{Supervised}}} \\
\multicolumn{1}{c|}{PCB\cite{sun2018beyond}} & \multicolumn{1}{c|}{ECCV'18} & 81.6 & 93.8 & 97.5 & \multicolumn{1}{c|}{98.5} & 69.2 & 83.3 & 90.5 & \multicolumn{1}{c|}{92.5} & 40.4 & 68.2 & - & - \\
\multicolumn{1}{c|}{DG-Net\cite{zheng2019joint}} & \multicolumn{1}{c|}{CVPR'19} & 86.0 & 94.8 & - & \multicolumn{1}{c|}{-} & 74.8 & 86.6 & - & \multicolumn{1}{c|}{-} & 52.3 & 77.2 & - & - \\ \hline
\end{tabular}%
}
\caption{Comparison with the state-of-the-art methods on Market1501, DukeMTMC-reID and MSMT17. The first and second best results among all unsupervised methods are, respectively, marked in \textcolor{red}{red} and \textcolor{blue}{blue}. $\dagger$ denotes using camera information. * denotes the backbone settings with IBN-Resnet and GeM pooling.}
\label{tab:sota}
\end{table*}

\subsection{Parameter Analysis}

\begin{figure}[t]
\centering
   \includegraphics[width=\linewidth]{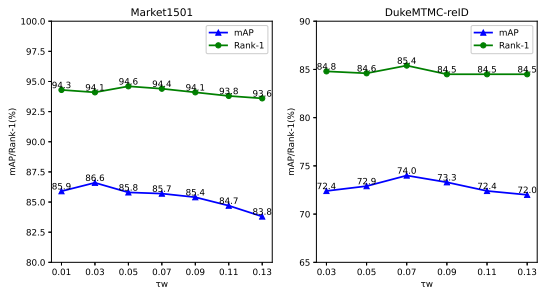}
   \caption{Parameter analysis of $\tau_w$ on Market1501 and DukeMTMC-reID.}
\label{fig:tauw}
\end{figure}

\begin{figure}[t]
\centering
   \includegraphics[width=\linewidth]{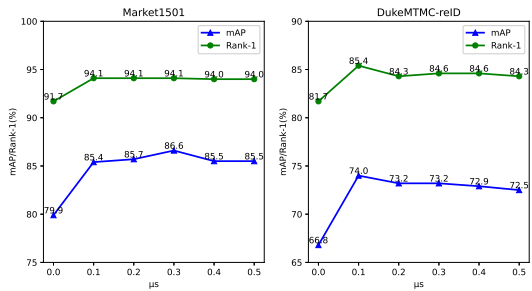}
   \caption{Parameter analysis of $\mu_s$ on Market1501 and DukeMTMC-reID.}
\label{fig:mu_s}
\end{figure}

We analyze the sensitivity of the hyper-parameter $\tau_w$ and $\mu_{s}$. The value of $\tau_w$ affects the weighting of the difficult instances in the cluster representation vectors. A larger value of $\tau_w$ means a larger proportion of difficult instances, but the cluster representation vectors will contain less global information of query instances. Conversely, the smaller the value of $\tau_w$ taken, the more the model will be biased towards locally optimal solutions and will not achieve higher performance. To find the optimal $\tau_w$ in Eq. \ref{eq:dyclloss}, we designed experiments to analyze the indicator curves of mAP and Rank-1 for $\tau_w$ ranging from 0.01 to 0.13 and intervals of 0.02, as shown in Figure \ref{fig:tauw}. The best results were obtained at $\tau_w=0.03$ on Market1501 and at $\tau_w=0.07$ on DukeMTMC-reID. In Eq. \ref{eq:smoothsoftlabel}, $\mu_s$ controls the significance of the redefined label in $L_{ss}$ for soft labels. We tune this parameter finely with the others fixed. Large $\mu_s$ will bias the model more towards eliminating errors due to image distortion, but the model will have more trouble in learning intra- and inter-class information. Small $\mu_s$ will result in a reduction in noise immunity. If $\mu_s = 0$, the loss function will decomposes down to be a cross-entropy loss function and cause the performance drop. Based on the experimental results in Figure \ref{fig:mu_s}, we set $\mu_s$ to 0.3 and 0.1 on Market1501 and DukeMTMC-reID, respectively.


\subsection{Comparison with State-of-the-Art Methods}

On three common datasets, \eg Market1501, DukeMTMC-reID, and MSMT17, we compared our proposed DCCC with various advanced USL Person Re-ID methods. The results are given in Tab. \ref{tab:sota}. It can be shown from the comparison that DCCC outperforms previous methods including LOMO\cite{liao2015person}, BOW\cite{zheng2015scalable}, BUC\cite{lin2019bottom}, MMCL\cite{wang2020unsupervised}, JVTC\cite{li2020joint}, JVTC+, HCT\cite{zeng2020hierarchical}, UGA\cite{wu2019unsupervised}, CycAs\cite{wang2020cycas}, SpCL\cite{ge2020self}, CAP\cite{wang2021camera}, ICE\cite{chen2021ice}, Cluster Contrast\cite{dai2022cluster}, and IIDS\cite{xuan2022intra}. Just like previous methods, DCCC do not use pre-trained data other than ImageNet. The baseline method Cluster Contrast utilizes clustering centroids too, but DCCC completely exploits the feature distribution of query instances. DCCC surpasses Cluster Contrast by 4.5\%/1.8\%, 1.4\%/0.5\%, 4.0\%/6.3\% for mAP/Rank1 on Market1501, DukeMTMC-reID and MSMT17. Moreover, unlike ICE, CAP and IIDS, DCCC don't use any camera information and achieves better performance with more limited information. Even more, DCCC outperformed some well-known supervised methods. Ultimately, our proposed method achieves 86.6\%/94.1\%, 74.0\%/85.4\%, 31.6\%/62.3\% results on Market1501, DukeMTMC-reID and MSMT17. Considering the different settings of backbone in ISE\cite{zhang2022implicit}, Cluster Contrast and HDCRL\cite{cheng2022hybrid}, we provide the results of DCCC with generalized-mean pooling (GeM) and IBN-Net for a fair comparison and discuss their effectiveness in the appendix.

\section{Conclusion}

In this paper, we propose a novel unsupervised Re-ID framework based on dynamic clustering and dynamic cluster contrastive learning. We design a dynamic cluster contrastive learning method with adaptive weights to store cluster representation vectors in cluster-level memory to solve the inconsistency problem. Then, we focus on the often overlooked clustering hyper-parameters by deploying a dynamic EPS scheduler to DBSCAN, resulting in a more stable clustering process. And we proposed a label smoothing soft contrastive loss to consider contrastive learning and self-supervised learning together with less computaional cost. Finally, we conduct experiments to validate the performance of the proposed method. Experiments' results show that our method has achieved the best performance comparing with those state-of-the-art methods. 

{\small
\bibliographystyle{unsrt}
\bibliography{egbib}
}

\clearpage

\appendix
\section{Appendix}
\subsection{Different Settings of Backbone}

IBN-Net\cite{pan2018two} uses instance normalization and batch normalization jointly on the basis of ResNet, which greatly improves the generalization and learning ability of the model, and solves the problem of domain transfer well on the Re-ID task, so we conducted related experiments using IBN-ResNet50-a. GeM\cite{radenovic2018fine} is capable of adaptively implementing feature space mapping, with a more robust feature representation compared to average pooling and maximum pooling. We also replaced the global average pooling (GAP)in the vanilla ResNet50 with generalized-mean pooling (GeM) for comparison. As shown in Tab. \ref{tab:ibngem}, both IBN-Net and GeM pooling gave a considerable improvement to the network.

\begin{table}[h]
\begin{center}
\resizebox{\linewidth}{!}{%
\begin{tabular}{c|cc|cc}
\hline
\multirow{2}{*}{Backbone} & \multicolumn{2}{c|}{Market1501} & \multicolumn{2}{c}{DukeMTMC-reID}  \\ \cline{2-5} 
             & mAP  & Rank-1 & mAP  & Rank-1 \\ \hline
Resnet50     & 86.6 & 94.1   & 73.5 & 85.5   \\
IBN-Resnet50 & 87.7 & 94.9   & 75.6 & 85.9   \\
Resnet50+GeM & 87.3 & 94.6   & 74.4 & 85.2   \\
\textbf{IBN-Resnet50+GeM}          & \textbf{88.2}  & \textbf{95.2}  & \textbf{76.9} & \textbf{87.1} \\ \hline
\end{tabular}%
}
\end{center}
\caption{Comparison of different settings of backbone on Market1501 and DukeMTMC-reID.}
\label{tab:ibngem}
\end{table}

\subsection{Comparison between $L_{ss}$ and $L_{ce}+L_{ss}$.}

Comparison tests are conducted to compare with the losses under the traditional single network configuration, and the findings are presented in Tab. \ref{tab:lss}. InfoNCE loss is a cross-entropy loss. displays the baseline cross-entropy loss ($L_{ce}$) based on a single st row network topology, while line 2nd row displays the situation in which $L_{ce}$ and $L_{ss}$ (+$L_{ss}$) are applied, with weights of 0.7. The experiments show that using $L_{ss}$ alone achieves better results with less number of parameters and computational effort than with the former.

\begin{table}[h]
\begin{center}
\resizebox{\linewidth}{!}{%
\begin{tabular}{c|cc|cc}
\hline
\multirow{2}{*}{Methods} & \multicolumn{2}{c|}{Market1501} & \multicolumn{2}{c}{DukeMTMC-reID}  \\ \cline{2-5} 
                         & mAP            & Rank-1         & mAP           & Rank-1        \\ \hline
baseline                 & 80.9           & 91.7           & 71.5          & 84.4          \\
+$L_{ss}$                     & 84.3           & 93.4           & 71.8          & 84.4 \\
$\mathbf{L_{ss}}$             & \textbf{84.8}  & \textbf{93.4}  & \textbf{72.5} & \textbf{84.5}          \\ \hline
\end{tabular}%
}
\end{center}
\caption{Comparison of different losses. The baseline method is based on cross-entropy loss ($L_{ce}$).}
\label{tab:lss}
\end{table}

\subsection{Sensitivity analysis of hyper-parameter $step$}

\begin{figure}[t]
\centering
   \includegraphics[width=\linewidth]{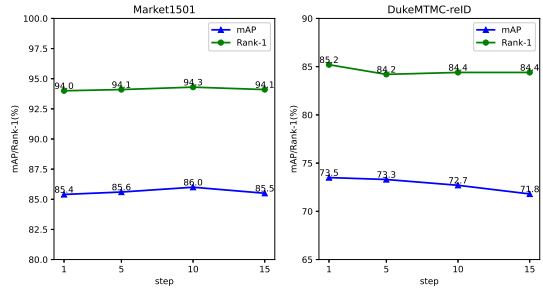}
   \caption{Parameter analysis of $step$ on Market1501 and DukeMTMC-reID.}
\label{fig:step}
\end{figure}

We validate the simplest step EPS scheduler to exploit how does it work like the step learing rate scheduler. We conducted experiments on the step EPS scheduler with $step=1,5,10,15$, and the experimental results are shown in Figure \ref{fig:step}. In particular, the step EPS scheduler degenerates to a linear EPS scheduler when $step=1$. It is easy to see that the performance degrades as $step$ grows for both the Market1501 and DukeMTMC-reID. Therefore, the linear EPS scheduler is better than the step EPS scheduler.

\end{document}